\documentclass[conference]{IEEEtran}
\IEEEoverridecommandlockouts
\usepackage{cite}
\usepackage{amsmath,amssymb,amsfonts}
\usepackage{algorithmic}
\usepackage{graphicx}
\usepackage{textcomp}
\usepackage{xcolor}

\usepackage{subcaption}

\usepackage{setspace}

\usepackage{tikz}
\usepackage{pgfplots}
\usepackage{pgfplotstable}
\usetikzlibrary{positioning, shapes, fit, calc, external, arrows, matrix}
\pgfplotsset{compat=1.15}

\tikzset{
    dot/.style   = {circle,          fill,      draw=black, minimum size=3pt, inner sep=0pt, outer sep=0pt},
    cross/.style = {cross out,       fill=none, draw=black, minimum size=4pt, inner sep=0pt, outer sep=0pt},
    rect/.style  = {rectangle,       fill,      draw=black, minimum size=3pt, inner sep=0pt, outer sep=0pt},
    tri/.style   = {regular polygon, fill,      draw=black, minimum size=5pt, inner sep=0pt, outer sep=0pt, regular polygon sides=3},
}

\DeclareMathOperator*{\argmin}{arg\,min}

\def\BibTeX{{\rm B\kern-.05em{\sc i\kern-.025em b}\kern-.08em
    T\kern-.1667em\lower.7ex\hbox{E}\kern-.125emX}}
\begin{document}

\title{A framework for compressing unstructured scientific data via serialization\\
\thanks{This manuscript has been authored in part by UT-Battelle, LLC, under
contract DE-AC05-00OR22725 with the US Department of Energy (DOE).
The publisher, by accepting the article for publication, acknowledges that
the U.S. Government retains a non-exclusive, paid up, irrevocable, world-
wide license to publish or reproduce the published form of the manuscript, or allow others to do so, for U.S. Government purposes. The DOE will provide
public access to these results in accordance with the DOE Public Access Plan
(http://energy.gov/downloads/doe-public-access-plan).}
}

\author{\IEEEauthorblockN{1\textsuperscript{st} Viktor Reshniak}
\IEEEauthorblockA{\textit{Computer Science and Math. Division} \\
\textit{Oak Ridge National Laboratory}\\
Oak Ridge, TN, USA \\
reshniakv@ornl.gov}
\and
\IEEEauthorblockN{2\textsuperscript{nd} Qian Gong}
\IEEEauthorblockA{\textit{Computer Science and Math. Division} \\
\textit{Oak Ridge National Laboratory}\\
Oak Ridge, TN, USA \\
gongq@ornl.gov}
\and
\IEEEauthorblockN{3\textsuperscript{rd} Rick Archibald}
\IEEEauthorblockA{\textit{Computer Science and Math. Division} \\
\textit{Oak Ridge National Laboratory}\\
Oak Ridge, TN, USA \\
archibaldrk@ornl.gov}
\and
\IEEEauthorblockN{4\textsuperscript{th} Scott Klasky}
\IEEEauthorblockA{\textit{Computer Science and Math. Division} \\
\textit{Oak Ridge National Laboratory}\\
Oak Ridge, TN, USA \\
klasky@ornl.gov}
\and
\IEEEauthorblockN{5\textsuperscript{th} Norbert Podhorszki}
\IEEEauthorblockA{\textit{Computer Science and Math. Division} \\
\textit{Oak Ridge National Laboratory}\\
Oak Ridge, TN, USA \\
pnb@ornl.gov}
}

\maketitle

\begin{abstract}
We present a general framework for compressing unstructured scientific data with known local connectivity. 
A common application is simulation data defined on arbitrary finite element meshes. 
The framework employs a greedy topology preserving reordering of original nodes which allows for seamless integration into existing data processing pipelines. 
This reordering process depends solely on mesh connectivity and can be performed offline for optimal efficiency.
However, the algorithm’s greedy nature also supports on-the-fly implementation.
The proposed method is compatible with any compression algorithm that leverages spatial correlations within the data. 
The effectiveness of this approach is demonstrated on a large-scale real dataset using several compression methods, including MGARD, SZ, and ZFP.
\end{abstract}


\section{Motivation and background}

The rapid advancement in computational prowess has notably propelled data production within scientific experiments and simulations. 
For example, a gas turbine jet engine simulation using the GE's GENESIS code requires billions of mesh nodes and thousands of time steps to resolve the complicated physics with sufficient accuracy \cite{ge-olcf}.
Each such simulation contains multiple double-precision variables resulting in data files easily exceeding tens of terabytes.
This creates a significant challenge for the storage, visualization and post-analysis of data due to the restrained capabilities of I/O and network systems \cite{Xie2012,Makrani2018}.
For instance, the Oak Ridge Leadership Computing Facility (OLCF) reported $8\times$ performance boost of their latest supercomputer Frontier~\cite{frontier} over its predecessor Summit~\cite{summit}.
Yet, improvements in storage and I/O only managed to achieve a meager $2-4\times$ enhancement.

For such data intensive applications, compression has emerged as an essential component of their processing pipelines enabling storage for later analysis.
At the same time, it introduces new "variables" into the problem raising concerns about the impact of compression on the validity of scientific data.
Lossless compression effectively eliminates this problem by providing the exact reconstruction of original data but is hardly useful due to essentially incompressible nature of floating-point scientific data \cite{SAYOOD20061,Zhao2020}.
Examples of lossless compressors include GZIP~\cite{deutsch1996gzip}, FPZIP \cite{Lindstrom2006}, FPC~\cite{burtscher2008fpc}, and ZSTD~\cite{collet2021rfc}.
At the same time, scientific data from simulations and experiments is often inherently lossy and can tolerate some error-controlled reduction in its accuracy.
Multiple approaches have been proposed to exploit this observation. 
The notable examples of lossy compressors are SZ \cite{tao2017significantly}, ZFP \cite{Lindstrom2014}, TTHRESH \cite{Ballester2020}, and ATC \cite{Baert2023}.

\begin{figure}[t]
    \centering
    \includegraphics[height=0.26\linewidth]{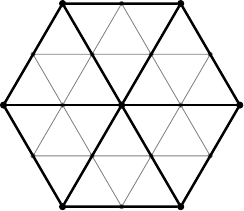}
    \includegraphics[height=0.26\linewidth]{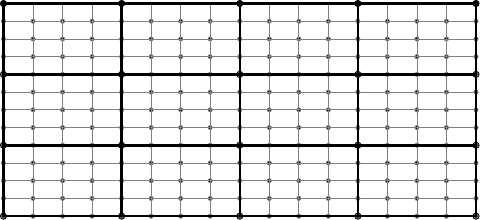}
    \caption{Dyadic structured grids supported by MGARD.}
    \label{fig:structured_grids}
\end{figure}

MGARD (MultiGrid Adaptive Reduction of Data) is a wavelet-like algorithm for compressing and refactoring scientific data based on the theory of multigrid
methods \cite{GONG2023101590,10254796,reshniak2024lifting}.
In MGARD, the compression is achieved via the hierarchical decomposition of data into the nested sequence of coarse projections and detail corrections.
This procedure decorrelates the original data and reduces it's entropy leading to better compressibility.
This is possible because the majority of scientific datasets have some level of spatial regularity meaning that the nodes that are close to each other are not expected to have drastically different values.
Besides it's excellent spatial decorrelation properties, the key advantage of MGARD
is its ability to a-priori control the compression error in various norms and derived quantities of interest~\cite{AinsworthMGARDQoI2019}.
As a possible limitation, the method in its current iteration can be only applied to structured data on arbitrary tensor-product grids and to certain simplicial meshes obtained via iterative refinement~\cite{AinsworthMGARDFE2020}, see Fig.~\ref{fig:structured_grids} for illustration.

\begin{figure}[t]
    \centering
    \begin{subfigure}[t]{\linewidth}
        \includegraphics[width=\linewidth]{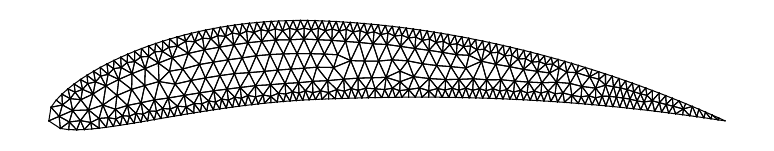}
        \caption{Unstructured mesh with $477$ nodes.}
        \label{fig:airfoil_mesh}
    \end{subfigure}
    \\[1em]
    \begin{subfigure}[t]{\linewidth}
        \includegraphics[width=0.55\linewidth]{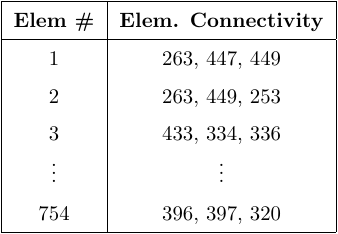}
        \;
        \includegraphics[width=0.39\linewidth]{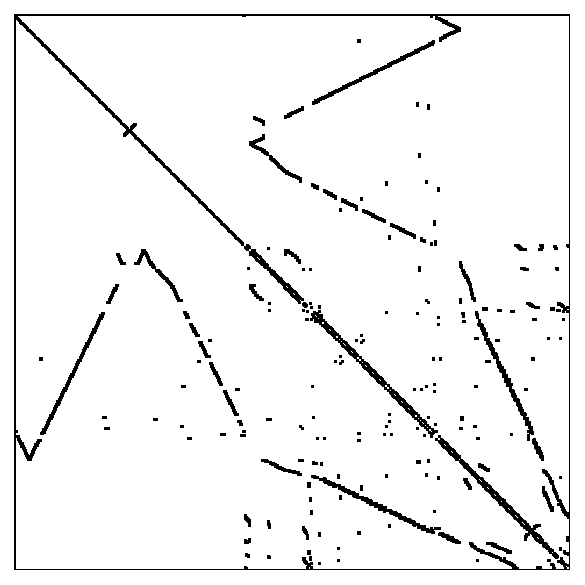}
        \caption{Element connectivity and adjacency matrix.}
        \label{fig:airfoil_connectivity}
    \end{subfigure}
    \caption{NACA9412 airfoil.}
    \label{fig:airfoil}

    \vspace{2em}
    
    \begin{subfigure}[t]{\linewidth}
        \includegraphics[width=\linewidth]{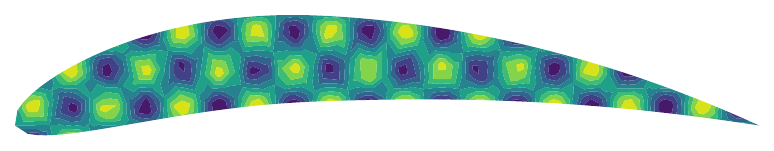}
    \end{subfigure}
    \\[1em]
    \begin{subfigure}[t]{\linewidth}
        \centering
        \includegraphics[width=0.9\linewidth]{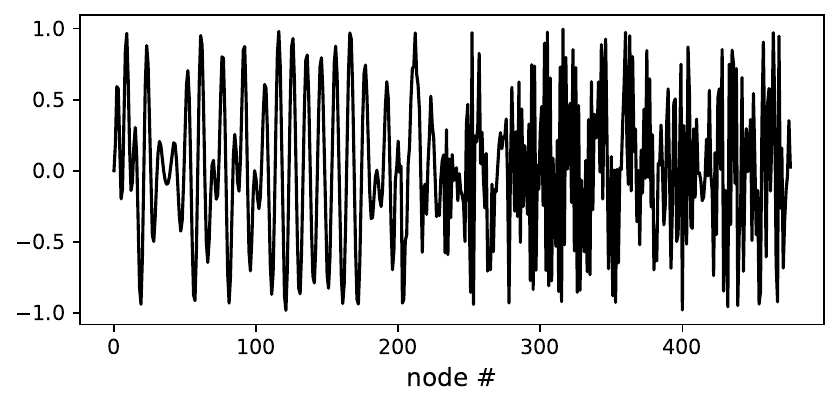}
    \end{subfigure}
    \caption{Smooth function over the NACA9412 airfoil and it's values at the flattened nodes with default ordering.}
    \label{fig:airfoil_function}
\end{figure}

Numerical simulations in complex geometries often restrict the applicability of structured approach and require more flexible solutions such as unstructured meshing \cite{BAKER200529}.
Fig.~\ref{fig:airfoil_mesh} gives a generic example of an unstructured mesh for the NACA9412 airfoil~\cite{enwiki:1237318857}. 
Unlike the structured grids in Fig.~\ref{fig:structured_grids}, unstructured meshes cannot be described by the nodes alone and require explicit specification of the nodal connectivity information.
Finite element meshes are commonly given as a collection of simply shaped elements, e.g., tetrahedrons or hexahedrons, each of which can be completely described by their nodes.
The CFD General Notation System (CGNS) defines useful conventions for representing such meshes \cite{CGNS}.
For example, a snapshot of the element connectivity for the grid in Fig.~\ref{fig:airfoil_mesh} is given in Fig.~\ref{fig:airfoil_connectivity} along its adjacency matrix.

The problem of devising a universal technique for directly compressing unstructured data is rather challenging due to the vastness of possible topological configurations of nodes and elements.
Of course, the trivial approach is to consider a one dimensional data representation following the global nodal indexing in the element connectivity.
Fig.\ref{fig:airfoil_function} demonstrates the apparent limitations of this idea.
Specifically, the default nodal ordering most certainly breaks the spatial correlations in data making it nearly random and hardly compressible.
Hence there is a need for an alternative way of representing unstructured data that can preserve the original correlations while converting it to the format amenable for compression with the available techniques.

To the best of our knowledge, no method has been proposed that could resolve the above issues without making certain compromises. 
A few recent efforts follow two lines of thoughts.
The first group of methods employ some form of mesh-to-grid data approximation. 
For example, \cite{el2021accurate,berger1991algorithm} explored converting the data into rectilinear grids to simplify analytic tasks such as visualization.
However, the error-controlled data reduction was not the primary focus of these works.
This task was considered in \cite{gong2024general} where the authors aimed to maintain the general structure of reduced data while mathematically preserving errors incurred during compression.
This was achieved by interpolating mesh data onto a rectilinear grid and then separately compressing interpolated data on the regular grid and one-dimensional residuals at the original nodes.
The method in \cite{gong2024general} demonstrated, on average, 2.3 -- 3.5x improvement in compression ratios when used with three state-of-the-art lossy compressors, within an error bound range of $10^{-6}-10^{-2}$.
As a possible limitation, this method requires one to keep the interpolation operator in the metadata for converting between the original and structured meshes.
It also involves a number of hyper-parameters that need to be tuned for the optimal performance.
However, the construction of the interpolation is purely geometry driven and can be performed offline before or during data collection.

The second group of methods involves reorganizing the original nodal data in a way that facilitates the compression task.
For example, the authors of \cite{Ren2024} proposed a prediction-traversal approach that sequentially visits/compresses mesh nodes until an unpredictable node is encountered.
Then the procedure is continued from a new seed until all nodes are visited.
This approach delivers improvements only within limited error ranges, requires storing traversal sequences, and has high computational overhead due to its dynamic nature.
For example, its demonstrated throughput is $<1$MB/s while most SOTA lossy compressors including MGARD achieve throughputs of tens to hundreds of MB/s \cite{GONG2023101590,zfp-rd100}.

\begin{figure}[t]
    \centering
    \includegraphics[height=0.7\linewidth]{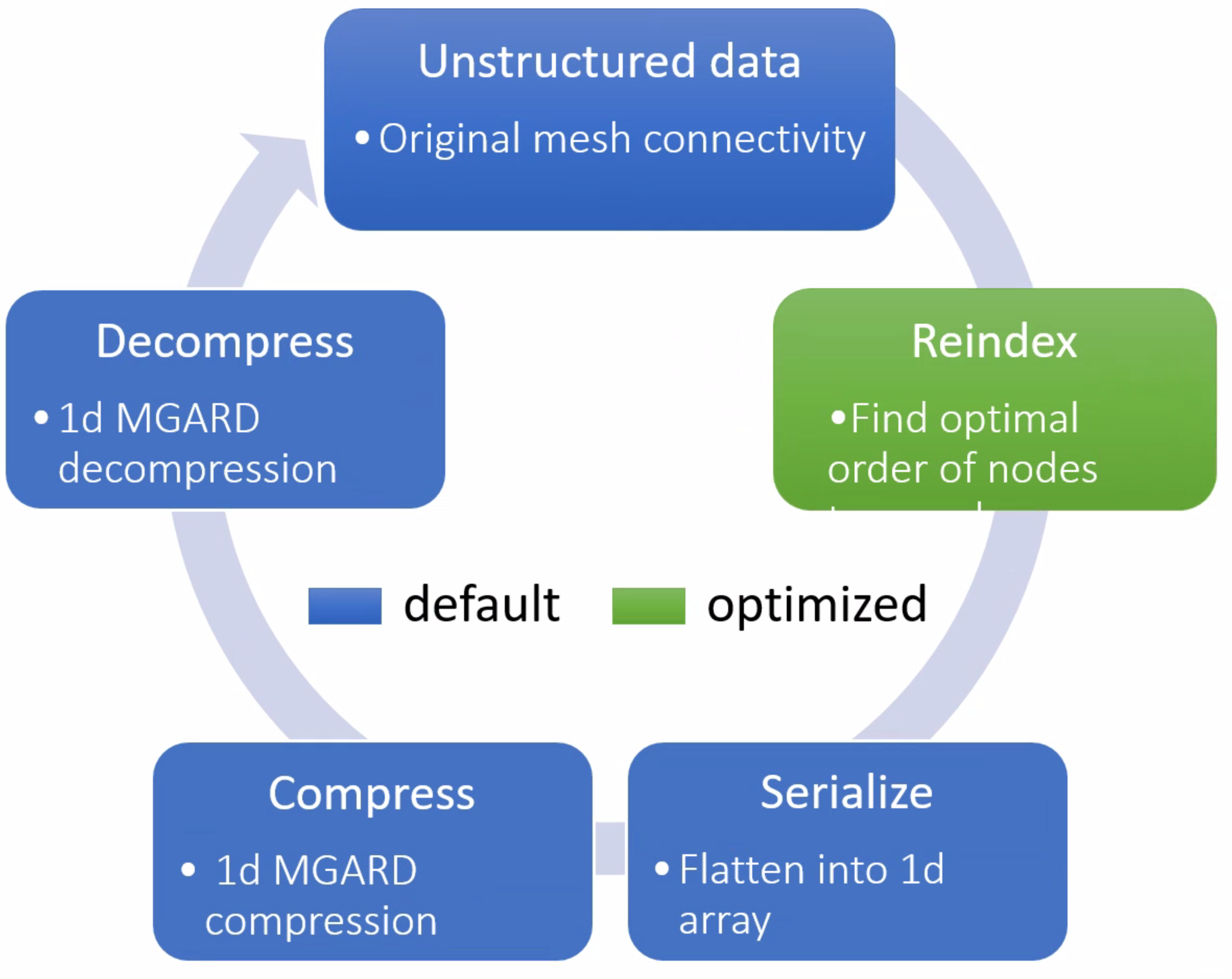}
    \caption{High level visualization of the proposed compression/decompression pipeline.}
    \label{fig:pipeline}
\end{figure}

Instead, in this effort we propose a node reordering technique that, similarly to the approach in \cite{gong2024general}, relies solely on the topological configuration of the original mesh and is deterministic.
The storage overhead of the proposed method is also low and involves storing a single integer array mapping the nodal indices.
The computational overhead involves one-time computation of the node reindexing that can be performed offline prior or during data generation.
Moreover, if the downstream tasks are not specific to the memory ordering of the grid nodes, storing the order-mapping array is not required.
In fact, most of the tasks in data processing pipelines involving unstructured grids are not specific to the particular node indexing.
The proposed method is remarkably simple and is illustrated schematically in Fig.~\ref{fig:pipeline}.

Before moving to the detailed description of the proposed algorithm in the next section, it is worth noting that the idea of choosing an optimal node indexing to improve performance is not new.
It is a widely recognized technique to reduce the storage and computation overhead of large scale sparse linear solvers used in finite element analysis.
The celebrated Cuthill-McKee and reverse Cuthill-McKee finite element basis ordering algorithms are the prominent examples of this idea \cite{Sherman1976,Fenves1983}.

\section{Algorithm}

Compressibility of grid data relies on its predictability.
If given the value at a node, one can exactly predict the values in some neighbourhood of the node, then the compression problem is solved completely.
In practice, one can predict the values with some finite degree of accuracy which relies on the regularity/smoothness of data.
However, even for very smooth data, it is possible to reliably predict values only at the nodes that are sufficiently close to each other.
This means that we need a way to find the node ordering that prefers close nodes to distant nodes.
The default node ordering in arbitrary unstructured grids does not fulfil this requirement in general.

At an abstract level, we want to solve a Minimum Linear Arrangement Problem (MinLA) which, given a graph $\mathbf{G}=(V,E)$ with nodes $V$ and edges $E$, aims at finding the mapping $\varphi:V\to\{1,\hdots,n\}$ onto a line topology such that the sum of the distances between adjacent nodes is minimized \cite{Petit2004}, i.e.,
\begin{align*}
    \varphi^* = \argmin_{\varphi} \sum_{uv\in E} |\varphi(u)-\varphi(v)|.
\end{align*}
MinLA is \textbf{NP}-hard and we certainly cannot hope to solve it in a reasonable time for graphs with node counts on the order of $10^6-10^9$.

\begin{figure}[tb]
\begin{algorithmic}[1]\onehalfspacing
    \renewcommand{\arraystretch}{2}
    \STATE \textbf{Given:} Graph $\mathbf{G}(V,E)$, node list $V$, edge list $E$
    \STATE \textbf{Output:} Nodal path $P$
    \STATE $P\gets \{\}$ \COMMENT{initialize list of visited nodes}
    \STATE $W\gets V$ \COMMENT{initialize list of unvisited nodes}
    \WHILE{$|W|>0$}
        \IF{$|\{v:uv\in E\}|>0$}
            \STATE $\displaystyle v^*\gets \argmin_{uv\in E} |uv|$ \COMMENT{closest connected node}
        \ELSE
            \STATE $v^*\gets \text{any } v\in W$ \COMMENT{any available node}
        \ENDIF
        \STATE $W\gets W \setminus \{v^*\}$
        \STATE $E\gets E \setminus \{uv^*\}$
        \STATE $P\gets P \cup \{v^*\}$
    \ENDWHILE
\end{algorithmic}
\caption{Greedy node indexing}\label{alg:greedy_indexing}
\vspace{2em}
\begin{algorithmic}[1]\onehalfspacing
    \renewcommand{\arraystretch}{2}
    \STATE \textbf{Given:} CGNS $ElementConnectivity$, node list $V$
    \STATE \textbf{Output:} Graph $\mathbf{G}=(V,E)$
    \STATE $E\gets \{\}$ \COMMENT{initialize edge list}
    \FOR{$e\in ElementConnectivity$}
        \FOR{$u,v\in e$}
            \IF{$u\neq v$}
                \STATE $E\gets E \cup \{uv\}$
            \ENDIF
        \ENDFOR
    \ENDFOR
\end{algorithmic}
\caption{Calculation of the traversal graph from the CGNS element connectivity.}\label{alg:adjacency_matrix}
\end{figure}

\begin{figure}[t]
    \centering
    \begin{subfigure}[t]{\linewidth}
        \includegraphics[width=0.45\linewidth]{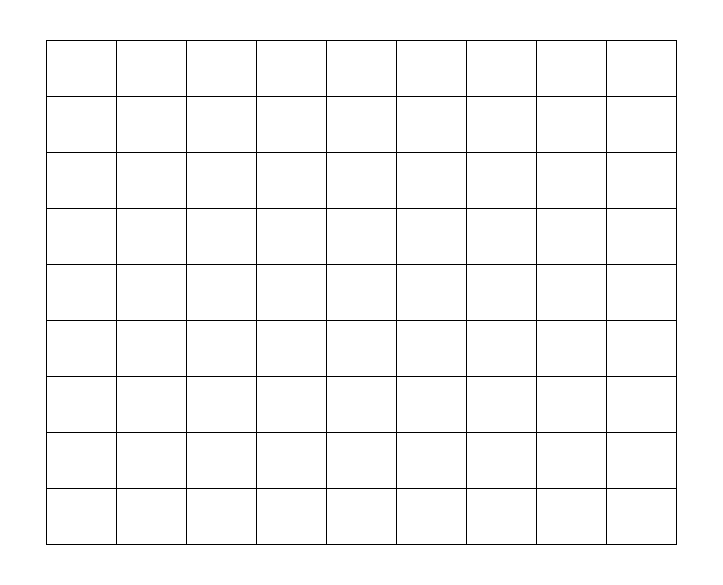}
        \includegraphics[width=0.45\linewidth]{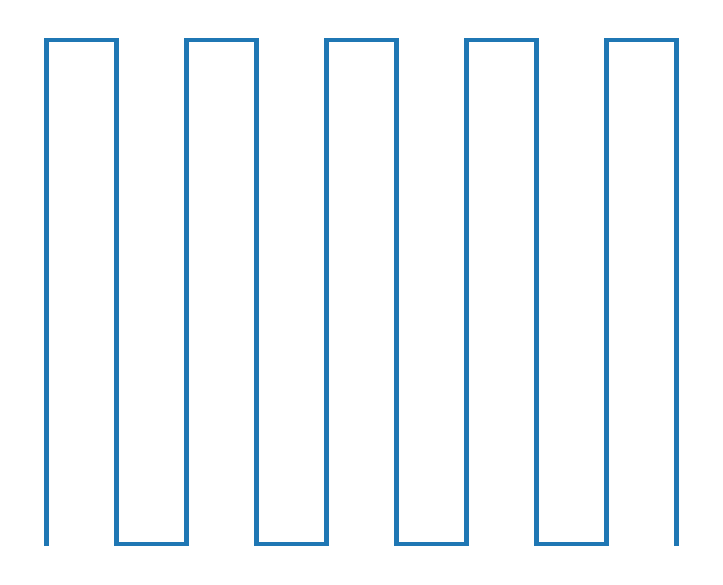}
    \end{subfigure}
    \begin{subfigure}[t]{\linewidth}
        \includegraphics[width=0.45\linewidth]{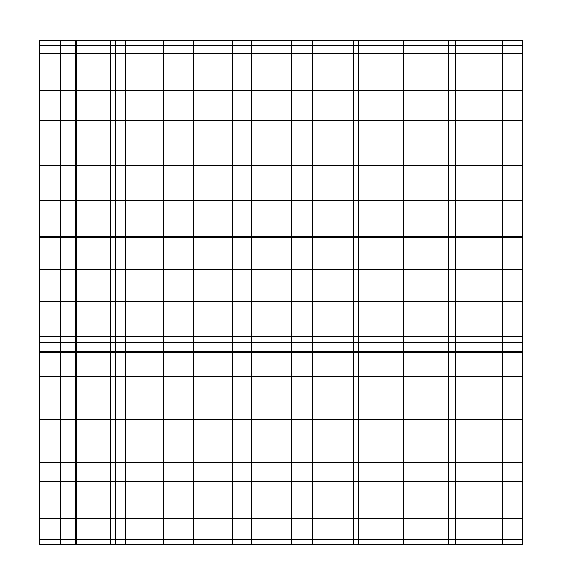}
        \includegraphics[width=0.45\linewidth]{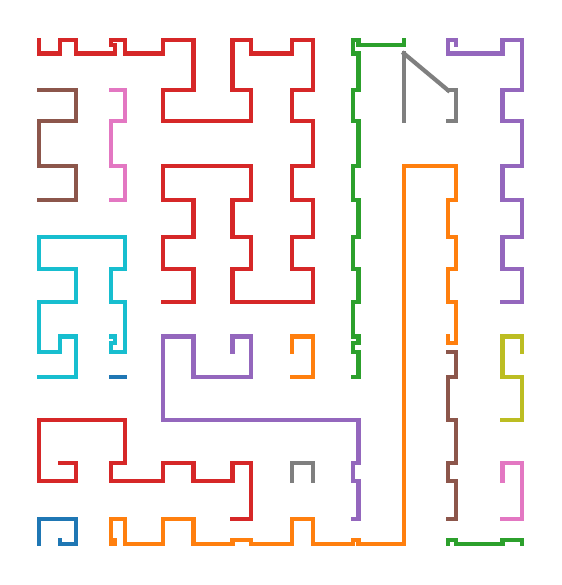}
    \end{subfigure}
    \caption{Structured meshes and greedy traversals.}
    \label{fig:box_path}
\end{figure}

\begin{figure}[t]
    \centering
    \begin{subfigure}[t]{\linewidth}
        \includegraphics[width=\linewidth]{images/airfoil.pdf}
        \includegraphics[width=\linewidth]{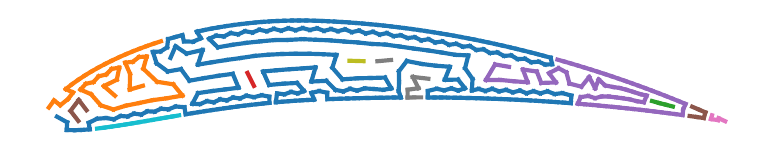}
    \end{subfigure}
    \\[1em]
    \begin{subfigure}[t]{\linewidth}
        \centering
        \includegraphics[width=0.9\linewidth]{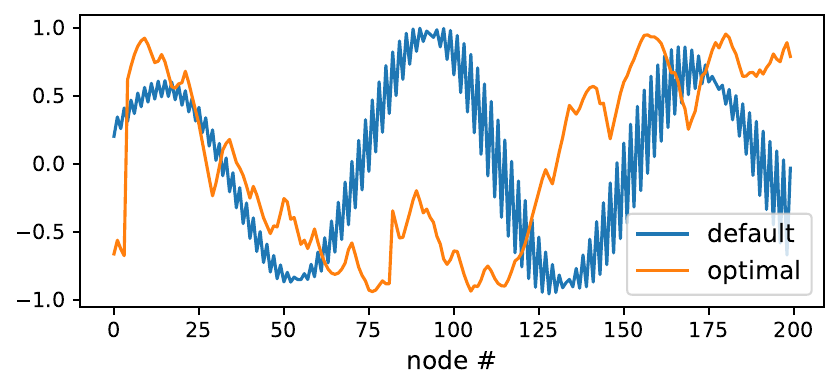}
    \end{subfigure}
    \caption{Original tetrahedral mesh of NACA9412 airfoil, optimal path over the mesh, and data values at the nodes over the default and optimal paths.}
    \label{fig:airfoil_path}
\end{figure}

Instead, we consider a relaxed greedy approximation described in Fig.\ref{alg:greedy_indexing}.
It starts with a first node of default ordering and proceeds in greedy fashion by selecting the closest unvisited node.
If the current node has no available connected neighbors, then an arbitrary node is chosen.
If the connectivity information is given in the form of CGNS Element Connectivity matrix as in Fig.~\ref{fig:airfoil_connectivity}, then the traversal graph $\mathbf{G}$ can be determined using Algorithm in Fig.~\ref{alg:adjacency_matrix}.
Note that in this case, the determined connectivity might be more permissive than the original mesh.
This happens, for instance, for quadrilateral meshes as in Fig.~\ref{fig:box_path}, which demonstrates the possibility of a traversal path with connections not in the original edge list (note the diagonal path).
For tetrahedral meshes, all the nodes of each element are connected to all the other nodes, and the above situation is impossible.
Fig.\ref{fig:box_path} also demonstrates that the traversal paths can have multiple disconnected components.
However, in higher dimensions, this is less likely.
Similarly, Fig.\ref{fig:airfoil_path} shows the traversal path over the simple tetrahedral mesh and the data values at the nodes along the default and greedy paths.
One can see that the ordered nodes result in much smoother profile as desired.

\section{Experiments}

In this section, we evaluate the proposed serialization framework using real-life data.
Specifically, we consider the dataset obtained from a simulation of the VKI (von Karman Institute) gas turbine blade cascade test case generated by GE’s GENESIS software \cite{arts1992aero}.
The dataset contains five double-precision variables defined on the predominantly quadrilateral grid with $21215679$ nodes.

To demonstrate the versatility of the proposed serialization approach, we consider three state-of-the-art lossy compressors summarized in Table~\ref{tab:compressor}.
All these compressors are specifically designed for compressing scientific data on rectilinear grids and require serialization of the unstructured data prior to the compression \cite{liang2020toward,liang2022toward}.

\newcolumntype{C}[1]{>{\centering\arraybackslash}p{#1}}
\begin{table}[h]
  \centering
  \renewcommand{\arraystretch}{1.5}
  \caption{Compressors used in our compression framework}
  {\small
  \begin{tabular}{C{1.4cm}|C{0.8cm}|C{4.99cm}}
    \hline
    Compressor & Version & GitHub repository \\
    \hline
    MGARD & 1.5.2 & https://github.com/CODARcode/MGARD\\
    \hline
    SZ3 & 3.1.8 & https://github.com/szcompressor/SZ3 \\
    \hline
    ZFP & 1.0.1 & https://github.com/LLNL/zfp \\
    \hline
  \end{tabular}
  }
  \label{tab:compressor}
\end{table}

To evaluate performance, we employ the following metrics: (1) Compression Ratio (CR), defined as the ratio of the original size to the reduced size, and (2) Relative Achieved Compression Error $\varepsilon$: the error measured in the reconstructed data below the specified tolerance $\tau$. 
The errors that we report are relative errors defined as $\varepsilon=\sqrt{\sum(x-x')^2}/\sqrt{\sum x^2}$, where the denominator is the $l_2$ norm of the original data.

\begin{figure}[t]
    \centering
    \includegraphics[width=\linewidth]{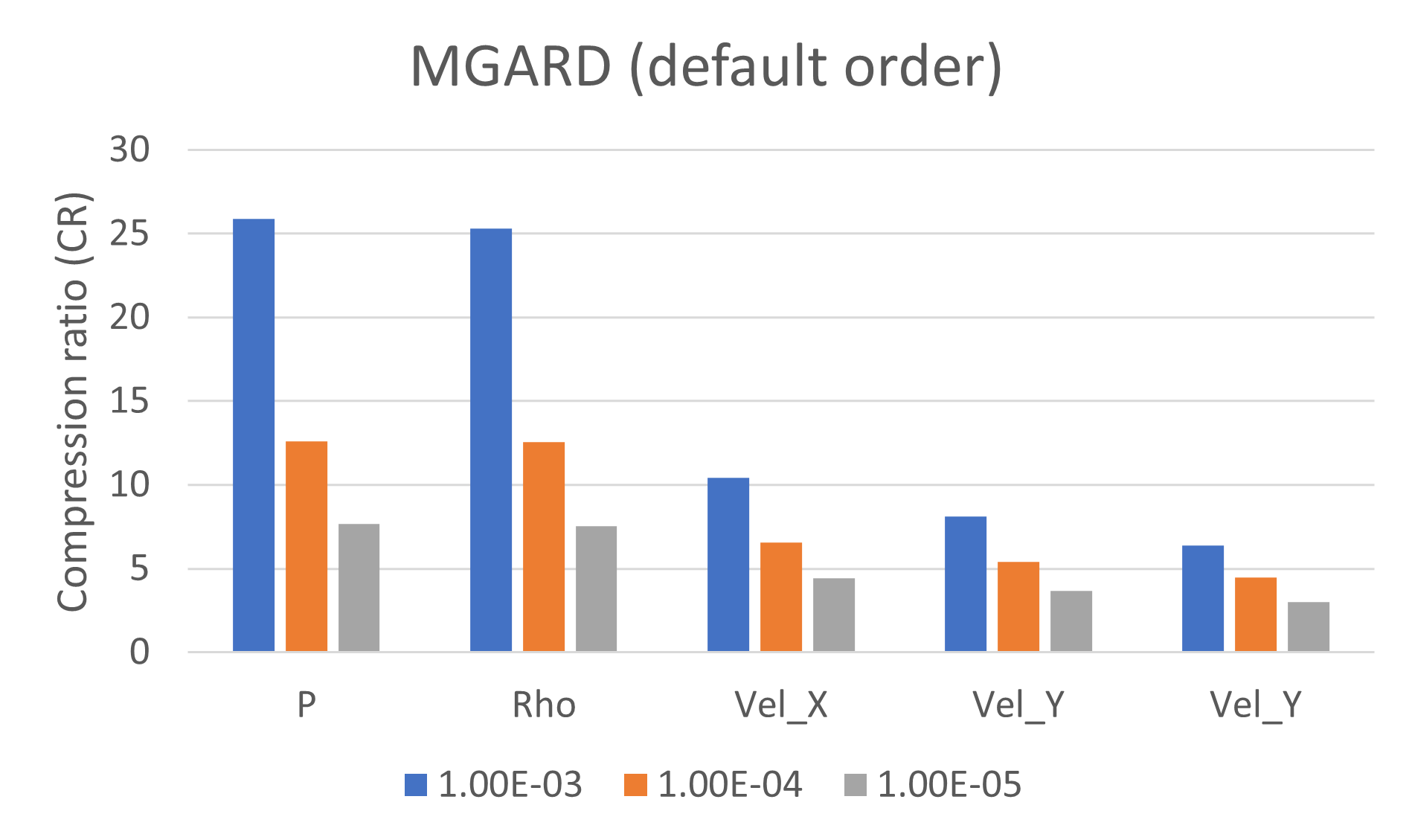}
    \includegraphics[width=\linewidth]{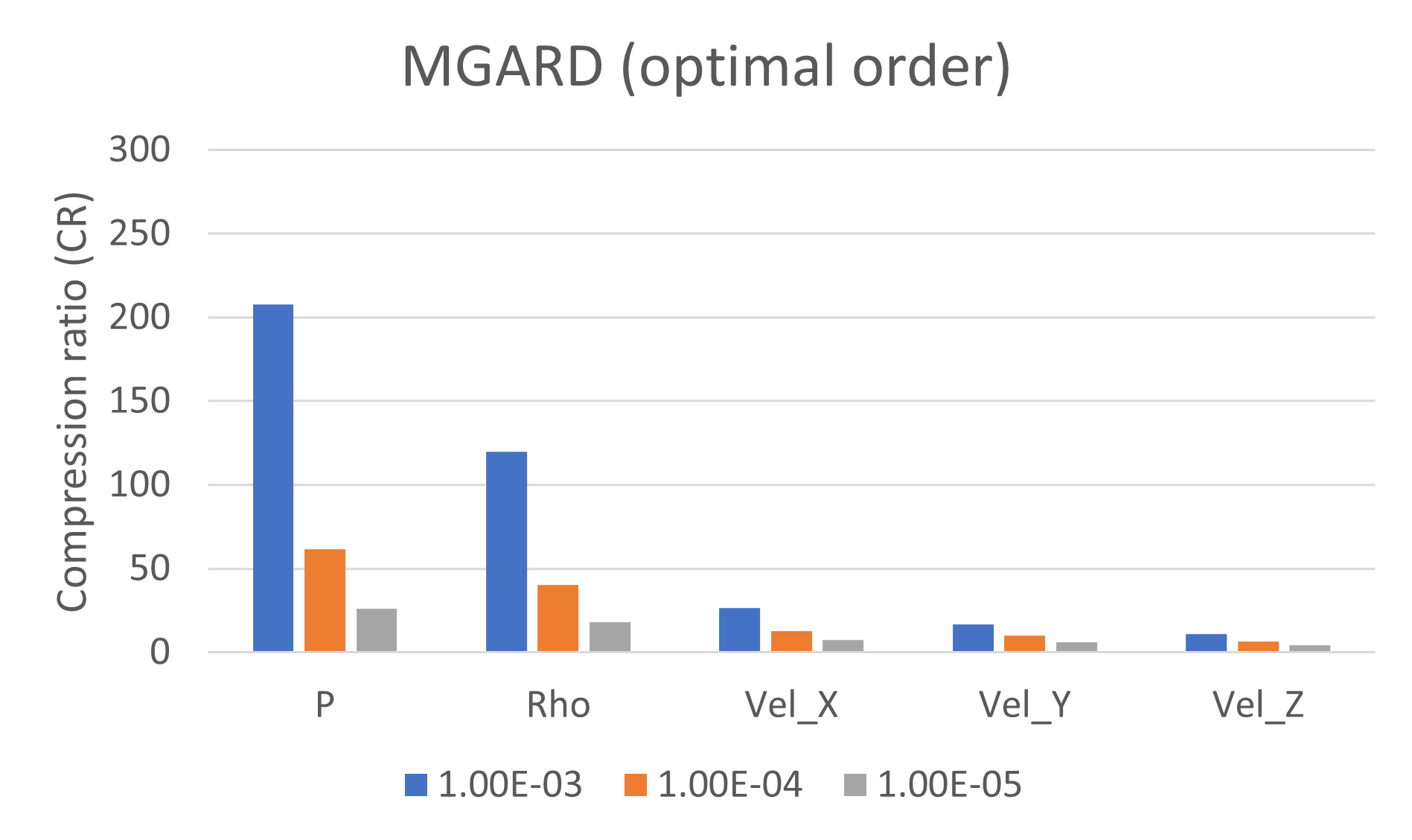}
    \includegraphics[width=\linewidth]{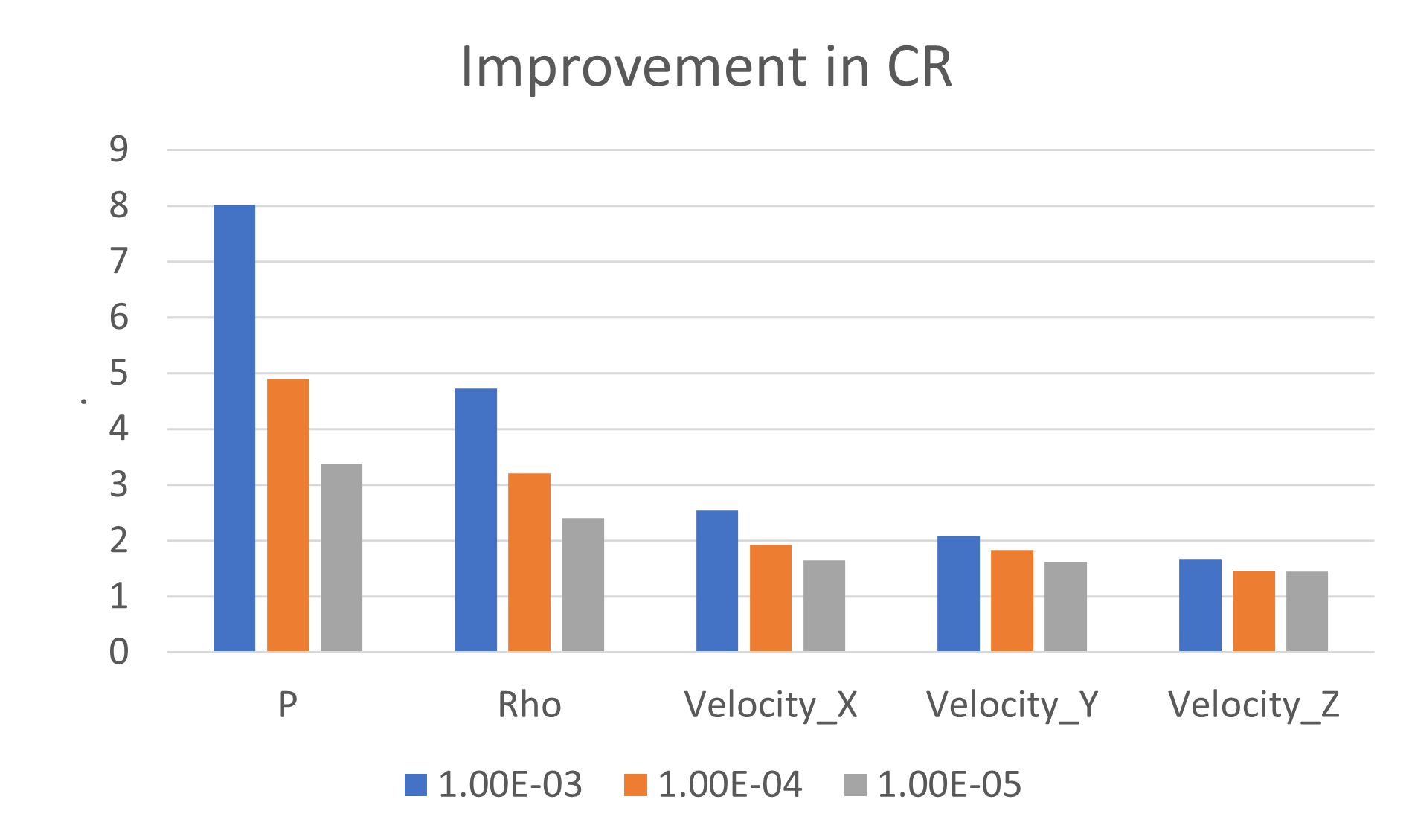}
    \caption{Compression ratios for the five variables in the VKI dataset and three values of the relative tolerance.}
    \label{fig:mgard_cr}
\end{figure}

Fig.~\ref{fig:mgard_cr} compares the compression ratios produced by MGARD for the default and optimized nodal orderings.
From the scale of the $y$-axis alone, it is immediately clear that the proposed optimization leads to the substantial improvement in the compressibility of data.
The largest improvement is observed for the pressure variable: 3.2x - 8x depending on the tolerance.
Indeed, pressure is the most regular variable of the dataset and is compressible well even for default ordering.
Contrary, velocities are the most "turbulent" variables meaning that the spatial correlation of their values is less pronounced.
This reflects in less impressive, yet still substantial, improvements in the compression ratios: 1.5x - 1.8x.
Another observable trend is the decay of compressibility improvements for smaller tolerances.
This outcome is anticipated because larger error bounds tend to quantize more residual data into zeros, resulting in greater improvements in overall compression ratios.

Finally, Table~\ref{tab:improvement} provides the summary of the observed total compression ratios for three evaluated compressors.
One can see, that both MGARD and SZ perform similarly with MGARD being marginally better.
Out of the three, ZFP compressor performed the worst which is due to the design of the transform coding approach employed by the method.
ZFP leverages data redundancies by dividing data into $4^d$ blocks and applying orthogonal transformations. 
Since the compression of residual data is performed in 1D instead of the $4\times4\times4$ space that the transformation algorithm favors, ZFP is likely to achieve less substantial improvement compared to the other two compressors.
Across the five variables and three error tolerances, the average improvement ratios for the MGARD, SZ, and ZFP methods are $2.2\times$, $2\times$, and $1.2\times$ respectively.

\begin{table}[]
    \centering
    \renewcommand{\arraystretch}{1.5}
    \caption{Observed total compression ratios for different values of the achieved relative tolerance.}
    \label{tab:improvement}
    \begin{tabular}{l|c|c|c}
    \hline
    Relative tolerance & $10^{-3}$ & $10^{-4}$ & $10^{-5}$ \\
    \hline
    \hline
    MGARD default      & 11.6 & 7.3 & 4.8 \\
    MGARD ordered      & 29.9 & 15  & 8.6 \\
    Improvement        & 2.6  & 2.1 & 1.8 \\
    \hline
    \hline
    SZ default         & 13   & 7.1  & 3.4 \\
    SZ ordered         & 28.7 & 13.6 & 6.9  \\
    Improvement        & 2.2  & 1.9  & 2.1 \\
    \hline
    \hline
    ZFP default        & 5.7 & 4.5 & 3.6 \\
    ZFP ordered        & 6.8 & 5.3 & 4.3 \\
    Improvement        & 1.2 & 1.2 & 1.2 \\
    \hline
    \end{tabular}
\end{table}

\section{Conclusion}

In this paper, we presented a data serialization compression framework specifically designed for scientific data on unstructured meshes, addressing the shortcomings of existing methods that mainly focus on rectilinear grid data. 
Our approach involves reindexing the mesh nodes to decrease the disorder in data values by exploiting known spatial correlations. 
Once the new traversal order is determined, it can be applied to the data in-place without changing existing data processing pipelines.
This solution is versatile and effective, easily integrating into current I/O and compression pipelines to improve the compression of unstructured mesh data with minimal changes to existing code. 
The tests permed on real-life dataset show that our method achieves, on average, a $1.2-2.2\times$ improvement in compression ratios when combined with three state-of-the-art lossy compressors, within an error bound range of $10^{-6}-10^{-2}$. 
Looking ahead, we plan to explore the benefits of using different ordering approaches and analyze their impact on throughput. 
Additionally, we will assess the overall time spent on compression and I/O of the reduced data.

\section*{Acknowledgment}

This research was supported by the SIRIUS-2 ASCR research project, the Scientific Discovery through Advanced Computing (SciDAC) program, specifically the RAPIDS-2 and FastMath SciDAC institutes, and the GE-ORNL CRADA data reduction project. 
This research used resources of the Oak Ridge Leadership Computing Facility, which is a DOE Office of Science User Facility.

\bibliographystyle{IEEEtran}
\bibliography{IEEEabrv,biblio}

\end{document}